# Logios : An open source Greek Polytonic Optical Character Recognition system

1st Konstantinos Perifanos
*National and Kapodistrian University of Athens)*
/ Athens, Greece
https://orcid.org/0000-0002-8352-8294

2nd Dionysis Goutsos
*National and Kapodistrian University of Athens*
Athens, Greece
https://orcid.org/0000-0002-2204-9740

*Abstract*—In this paper, we present an Optical Character Recognition (OCR) system specifically designed for the accurate recognition and digitization of Greek polytonic texts. By leveraging the combined strengths of convolutional layers for feature extraction and recurrent layers for sequence learning, our system addresses the unique challenges posed by Greek polytonic scripts. This approach aims to overcome the limitations of traditional OCR methods, offering significant improvements in accuracy and efficiency. We release the underlying model as an open-source library and make our OCR platform available for academic use.

## I. Introduction

Historical Greek polytonic scripts have a rather complex target vocabulary and various set of rules resulting in a large character set of more than 200 characters, including the acute accent, the grave accent, the circumflex, the rough breathing (dasi pneuma), the smooth breathing (psilon pneuma), the diaeresis and the iota subscript [1].

At the same time, open-source resources for optical character recognition in historical polytonic texts are, for the most part, limited, outdated, suboptimal, and/or very complicated to work with.

In this paper, we present our work aimed at addressing both the performance issues and the complexity and availability of existing open-source tools.

More concretely, our contribution is three-fold:
- We extend the work of [1] by using a modified Convolutional Recurrent Neural Networks (CRNN) architecture [2] in the place of LSTMs and we report a significant improvement in the evaluation results, effectively reducing the character error rate to approximately 1.18% on average on their publicly available dataset
- We release an additional dataset consisting of 6796 training examples
- We are open sourcing our best models and data into an easy to use Python library and we present Logios, an OCR tool for polytonic texts.[1]

## II. Related Work

Optical Character Recognition of historical documents is a very well studied problem. Simistira et al. [1] are using bidirectional LSTM with CTC loss. Katsouros [14] are applying Hidden Markov Models on degraded historical documents. Roberson [15] are applying OCR in a large collection of 1,200 volumes comprising of approximately 329 million words. Sichani et. al [16] are presenting a framework for development, optimization and quality control in OCR systems.

With the advent of Deep Learning, frameworks such as PyTorch [3], the definition of complex neural network architectures became much easier. At the same time, recent advances in deep learning optimization techniques, beyond the traditional Stochastic Gradient Descent (SGD), such as AdamW [7] as well as the literature on normalization layers such as BatchNorm, GroupNorm and LayerNorm [4]–[6] resulted in much more stable and effective training of potentially very deep neural network architectures. Layer normalization techniques are typically employed in deep neural networks to stabilize training and enhance performance. BatchNorm normalizes activations across a batch of samples for each feature, addressing issues such as vanishing/exploding gradients. LayerNorm normalizes across all features within a single sample, making it suitable for recurrent networks and transformers. GroupNorm offers a compromise by dividing channels into groups and normalizing within each, balancing stability with less reliance on batch size.

In this work, we are leveraging these advances along with standard practices followed in recent

---
[1]https://logios.phil.uoa.gr or http://kalchasocr.phil.uoa.gr/logios

Deep Learning literature, and we report significant improvement in the performance of the resulting model.

## III. Dataset

As a starting point, we trained our models using the Polyton-DB[2] dataset released by [1] (See Table I for Polyton DB details).

After obtaining a high-quality neural network trained on the aforementioned dataset, we used it to annotate and label additional documents, resulting in an extended collection of 6,796 additional training examples (text lines). Note that the actual number of examples in our labeled dataset is much larger; however, we restricted the selection to those containing only characters present in the Polyton-DB dataset. That is, we omitted examples with digits, Latin script, etc., to maintain consistency between the datasets and the resulting models.

The details of this dataset can be found in table II.

TABLE I
Details about various datasets in the Polyton-DB.

| Set | Pages | Textlines |
|---|---|---|
| Greek Parliament Proceedings | | |
| Vlahou | 4 | 373 |
| Markezinis | 18 | 1,666 |
| Saripolos | 6 | 642 |
| Venizelos | 5 | 522 |
| Greek Official Government Gazette | 5 | 687 |
| Appian's Roman History | 315 | 11,799 |
| Synthetic data | 315 | 11,799 |
| Total | 353 | 15,689 |

TABLE II
Kalchas dataset

| Type | Textlines |
|---|---|
| Academic (Humanities) | 1465 |
| Academic (Natural Sciences) | 832 |
| Academic (Social Sciences) | 853 |
| Religious | 465 |
| Speech (Other) | 1440 |
| Total | 6796 |

## IV. Model Architecture and Training

The underlying Kalchas OCR model is based on a CRNN architecture with CTC loss, especially suitable for tasks where the output sequence is a transcription of the input. The network takes as input the input image (textline) and produces a sequence of characters (the recognized text). We are using a modified version of the CRNN architecture as introduced in [2]. More concretely, we are replacing the BatchNorm layer normalization with GroupNorm layer normalization, essentially decoupling the learning process dependency on batch statistics. The effect of layer normalization generally leads to a more stable training process and also in our experiments, to better performance, by reducing the character error from an initial 2% using the vanilla CRNN, to 1.18% in the Polyton-DB dataset.

### A. Preprocessing

All training images are binarized and resized to 760x80 pixels using bicubic interpolation. In our experiments, the choice of the interpolation strategy affects the quality of the model. Bicubic interpolation produces the best results in the same train/test split. No global (page level) denoising steps are performed before binarization and segmentation, although this can generally lead to potentially better results in heavily degraded images.

### B. Training

We optimized the model using Connectionist Temporal Classification Loss (CTC Loss) [8]. CTC is a natural loss function choice in problems where the input and the output consists of variable-length sequences. It addresses the challenge of aligning these variable-length input sequences with output labels. CTC allows the network to output a probability distribution at each time step, including a "blank" symbol. During training, the most likely sequence of labels is determined through a decoding process, enabling the network to learn the mapping between input and output sequences without explicit alignment information.

We are optimizing the network using the RMSProp optimizer [9] for 100 epochs, with batch size 32. We trained and we are releasing two different models, a model based on the Polyton-DB dataset and a second model trained both in Polyton-DB and Kalchas dataset.

The model training process is rather straightforward, as we used a MacBook M3 Pro train the model for 100 epochs, for a total training duration of approximately 24 hours for the first model and 36 hours for the second. We train both models on a 90% split of the data and evaluate it on the rest 10%.

In inference time, the network is able to process 7 inputs/sec on a CPU and approximately 14 inputs/sec on an MPS device.

[2] http://media.ilsp.gr/PolytonDB/

## V. Results

In this section we present and discuss the performance of the model and how it responds to out-of distribution inputs and most common mistakes. As expected, the model performs very well in documents with similar fonts and quality, as in the training data.

Our best model trained on Polyton-DB achieves Character Error Rate (CER) 1.18% and Word Error Rate (WER) 0.76%.[3]. Additionally, we performed evaluation on the validation dataset using the Tesseract engine [17], which also supports polytonic Greek. It is worth noting that the most recent Tesseract version we used in our evaluations[4] is significantly improved compared to the results reported by [1]. More specifically, we find that the character error obtained by tesseract in our validation split has been reduced to 9.7% (from 71.43% in [1]) and the word error rate to 21.26% from 71.43%) correspondingly.

See table V-A for a comparison of the different model performances.

We are also releasing a second model, trained both on Polyton-DB and on 6796 additional training examples from the Kalchas dataset. This model achieves 1.32% Character Error Rate and 0.9% Word Error Rate. Choosing and using a model from the Kalchas library is straightforward:

```python
from PIL import Image
from kalchas.ocr import list_available_models,\
            load_ocr_model

#get all available models: ['Polyton-DB','Kalchas']
models = list_available_models()
model = load_ocr_model('Kalchas')

image_path = "images/010000.bin.png"
image  = Image.open(image_path).convert('L')

text = model.ocr([image])
```

Listing 1. Kalchas example

### A. Examples

In the following, we present examples and outputs of the models in the validation set, as well as in data outside the training/test splits.

TABLE III
Character Error Rate Comparison

| OCR Engine | Character Error Rate (%) | Word Error Rate |
|---|---|---|
| Simistira et. al | 5.51 | 24.13 |
| Tesseract | 9.7 | 21.26 |
| **Kalchas** | **1.12** | **0.76** |

---

[3]The code used to train the model and the corresponding data splits are available at http://github.com/kperi/Kalchas
[4]tesseract 5.4.1, leptonica-1.84.1

Fig. 1. Example text line from the test set, recognized with 100% accuracy

προσεγίγνετο. ἦν τε διὰ στόματος ἐπὶ πᾶσιν

Fig. 2. Recognition failure example. The segment is recognised as '60'

208

Fig. 3. Out of sample examples, taken from a document outside the training set

τὸ ἡγεμονικὸν τῆς ψυχῆς του προσηλωμένον εἰς θεωρίας ψυχήν τουνὰ ἔχηην καὶ τὸ ββασιλικὸνσαῆπτρονεῖς τὰοχεῖράς.

An interesting observation here is that the recognition quality is heavily dependent on the quality of the segmentation. For example, if segmentation is not optimal and residuals of previous or next lines appear in a given segment, the recognition accuracy drops significantly. This can be mitigated with data augmentation techniques, where artificial noise can be added to inputs to increase the robustness of the inference and the overall stability of the model.

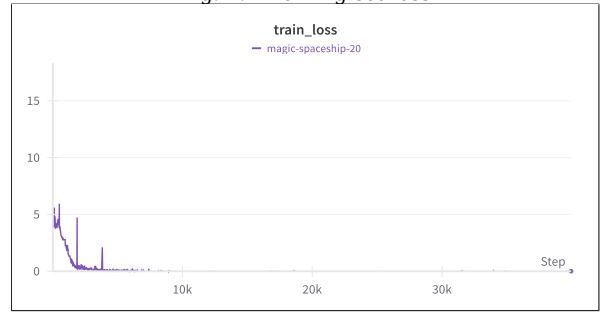

Fig. 4. Training set loss

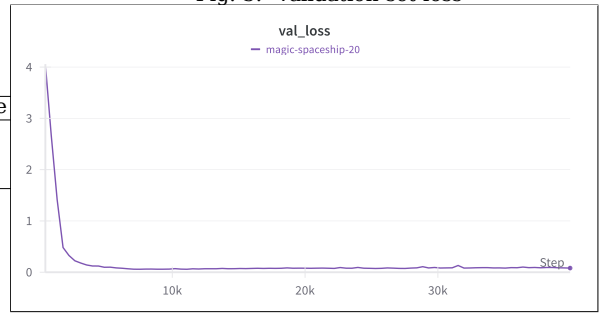

Fig. 5. Validation set loss

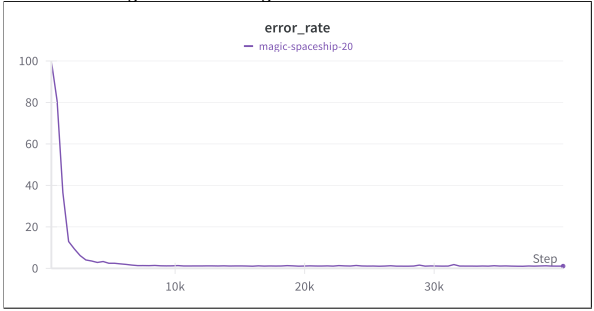

Fig. 6. Training set Character Error Rate

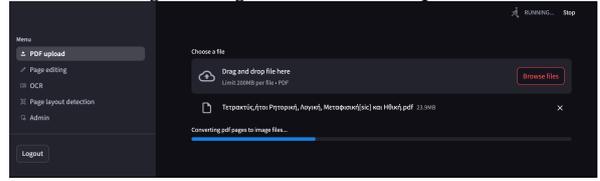

Fig. 7. Logios - Document upload

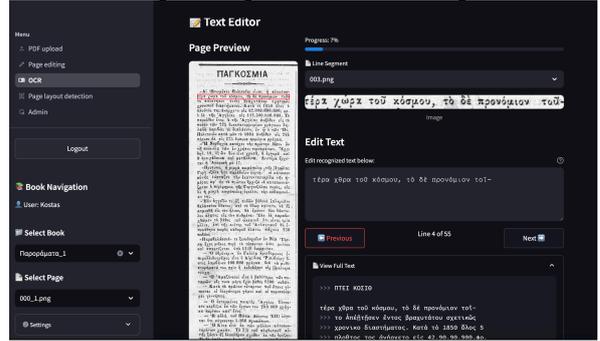

Fig. 8. Logios: OCR and editing

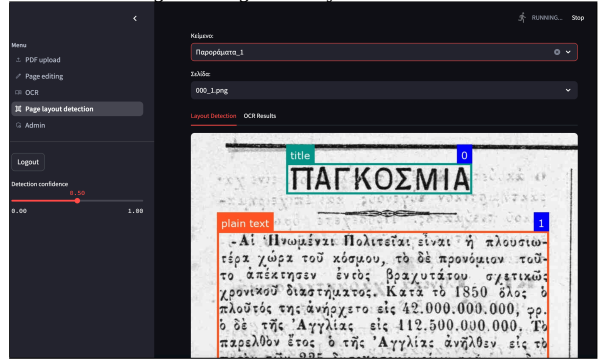

Fig. 9. Logios - Layout detection

| Pair 1 | Pair 2 | Count |
|---|---|---|
| ὐ | ὐ | 3 |
| η | η | 3 |
| . | , | 3 |
| π | τ | 2 |
| , | . | 2 |
| α | ε | 2 |
| χ | γ | 2 |
| ά | ά | 2 |
| τ | π | 2 |
| β | ό | 2 |
| χ | ν | 2 |

TABLE IV
Table of Character Pairs and Counts

## VI. Logios, a Polytonic OCR platform

To facilitate training and testing, we developed a dedicated application for document uploading, segmentation, OCR, and label collection. Originally designed to bootstrap our polytonic corpora-building tasks, we quickly realized that such a tool would be highly valuable to the academic community. As a result, we decided to further improve the application and offer it as a free tool to the community.

Logios is implemented using the Kalchas OCR library and Python's Streamlit library as the UI engine. We use OpenCV for image processing, Kraken for document denoising and line segmentation, and DocLayout-YOLO for layout analysis.

Our system supports page extraction from PDF images, as well as both manual and automatic page layout detection and optical character recognition. The produced OCR-ed text can be edited accordingly, allowing our tool to be used for document labeling. Additionally, we offer the ability to train or fine-tune our existing models on historical documents that deviate from those originally used to train Kalchas.

## VII. Limitations and Future work

The work presented here is based on a combination of recurrent and convolutional neural networks, leveraging both the sequential and spatial nature of the input domain. Recent work [12] emphasizes end-to-end systems capable of simultaneously performing page layout analysis, optical character recognition, and text scene detection [13].

We aim to extend this work by incorporating page layout detection and exploring the latest state-of-the-art architectures, such as Transformers [10] and Visual Transformers (ViT) [11]. Additionally, data augmentation techniques were not used in this work. Polyton-DB is already preprocessed, and our dataset was created using standard existing techniques for line segmentation and de-skewing. We also experimented with document denoising as a preprocessing step before the segmentation phase, and we aim to refine the corresponding models.

## References

[1] Simistira, F., Ul-Hassan, A., Papavassiliou, V., Gatos, B., Katsouros, V. & Liwicki, M. Recognition of historical Greek polytonic scripts using LSTM networks. *2015 13th International Conference On Document Analysis And Recognition (ICDAR)*. pp. 766-770 (2015)

[2] Shi, B., Bai, X. & Yao, C. An End-to-End Trainable Neural Network for Image-based Sequence Recognition and Its Application to Scene Text Recognition. (2015), https://arxiv.org/abs/1507.05717

[3] Paszke, A., Gross, S., Chintala, S., Chanan, G., Yang, E., DeVito, Z., Lin, Z., Desmaison, A., Antiga, L. & Lerer, A. Automatic differentiation in PyTorch. (2017)

[4] Ioffe, S. & Szegedy, C. Batch Normalization: Accelerating Deep Network Training by Reducing Internal Covariate Shift. (2015), https://arxiv.org/abs/1502.03167

[5] Wu, Y. & He, K. Group Normalization. (2018), https://arxiv.org/abs/1803.08494

[6] Ba, J., Kiros, J. & Hinton, G. Layer Normalization. (2016), https://arxiv.org/abs/1607.06450

[7] Loshchilov, I. & Hutter, F. Decoupled Weight Decay Regularization. (2019), https://arxiv.org/abs/1711.05101

[8] Graves, A. & Graves, A. Connectionist temporal classification. *Supervised Sequence Labelling With Recurrent Neural Networks*. pp. 61-93 (2012)

[9] Tieleman, T. Lecture 6.5-rmsprop: Divide the gradient by a running average of its recent magnitude. *COURSERA: Neural Networks For Machine Learning*. **4**, 26 (2012)

[10] Vaswani, A., Shazeer, N., Parmar, N., Uszkoreit, J., Jones, L., Gomez, A., Kaiser, L. & Polosukhin, I. Attention Is All You Need. (2023), https://arxiv.org/abs/1706.03762

[11] Dosovitskiy, A., Beyer, L., Kolesnikov, A., Weissenborn, D., Zhai, X., Unterthiner, T., Dehghani, M., Minderer, M., Heigold, G., Gelly, S., Uszkoreit, J. & Houlsby, N. An Image is Worth 16x16 Words: Transformers for Image Recognition at Scale. (2021), https://arxiv.org/abs/2010.11929

[12] Zhu, W., Sokhandan, N., Yang, G., Martin, S. & Sathyanarayana, S. DocBed: A multi-stage OCR solution for documents with complex layouts. *Proceedings Of The AAAI Conference On Artificial Intelligence*. **36**, 12643-12649 (2022)

[13] Long, S., Qin, S., Panteleev, D., Bissacco, A., Fujii, Y. & Raptis, M. Towards end-to-end unified scene text detection and layout analysis. *Proceedings Of The IEEE/CVF Conference On Computer Vision And Pattern Recognition*. pp. 1049-1059 (2022)

[14] Katsouros, V., Papavassiliou, V., Simistira, F. & Gatos, B. Recognition of Greek polytonic on historical degraded texts using HMMs. *2016 12th IAPR Workshop On Document Analysis Systems (DAS)*. pp. 346-351 (2016)

[15] Robertson, B. & Boschetti, F. Large-scale optical character recognition of ancient greek. *Mouseion*. **14**, 341-359 (2017)

[16] Sichani, A., Kaddas, P., Mikros, G. & Gatos, B. OCR for Greek polytonic (multi accent) historical printed documents: development, optimization and quality control. *Proceedings Of The 3rd International Conference On Digital Access To Textual Cultural Heritage*. pp. 9-13 (2019)

[17] Smith, R. An overview of the Tesseract OCR engine. *Ninth International Conference On Document Analysis And Recognition (ICDAR 2007)*. **2** pp. 629-633 (2007)